\documentclass[preprint,12pt,sort&compress]{elsarticle}

\usepackage{amssymb}
\usepackage{amsmath}

\journal{}

\usepackage{graphicx}
\usepackage{subcaption}
\usepackage{tabularx}
\usepackage{array}          
\usepackage{booktabs}   
\usepackage{makecell}
\usepackage{enumitem}
\usepackage{hyperref}
\usepackage{lineno}
\usepackage{xcolor}

\begin{document}

\begin{frontmatter}



\title{Deeply-Conditioned Image Compression via Self-Generated Priors}

\affiliation[sustech]{organization={College of Engineering, Southern University of Science and Technology},
                     city={Shenzhen},
                     country={China}}

\affiliation[pcl]{organization={Peng Cheng Laboratory},
                  city={Shenzhen},
                  country={China}
                  }
\affiliation[pek]{organization={Peking University}
                  }
                  
\author[sustech,pcl]{Zhineng Zhao}

\author[sustech,pcl]{Zhihai He\corref{cor1}}
\ead{hezh@sustech.edu.cn} 
\cortext[cor1]{Corresponding author} 
\author[pcl]{Zikun Zhou}
\author[pek]{Siwei Ma}
\author[pcl]{and Yaowei Wang}

\begin{abstract}
Learned image compression (LIC) has shown great promise for achieving high rate-distortion performance. However, current LIC methods are often limited in their capability to model the complex correlation structures inherent in natural images, particularly the entanglement of invariant global structures with transient local textures within a single monolithic representation. This limitation precipitates severe geometric deformation at low bitrates. To address this, we introduce a framework predicated on functional decomposition, which we term Deeply-Conditioned Image Compression via self-generated priors (DCIC-sgp).
Our central idea is to first encode a potent, self-generated prior to encapsulate the image's structural backbone. This prior is subsequently utilized not as mere side-information, but to holistically modulate the entire compression pipeline. This deep conditioning, most critically of the analysis transform, liberates it to dedicate its representational capacity to the residual, high-entropy details. 
This hierarchical, dependency-driven approach achieves an effective disentanglement of information streams. 
Our extensive experiments validate this assertion; visual analysis demonstrates that our method substantially mitigates the geometric deformation artifacts that plague conventional codecs at low bitrates. Quantitatively, our framework establishes highly competitive performance, achieving significant BD-rate reductions of 14.4\%, 15.7\%, and 15.1\% against the VVC test model VTM-12.1 on the Kodak, CLIC, and Tecnick datasets.


\end{abstract}



\end{frontmatter}


\section{Introduction}
\label{sec:intro}
The exponential growth of visual data has escalated the demand for efficient image compression. Traditional codecs such as JPEG~\cite{wallace1991jpeg}, WebP~\cite{google2010webp}, and VVC~\cite{team2021vvc} struggle to maintain visual quality at low bitrates, particularly when processing complex image content with heterogeneous textures and geometric structures. 
Learned image compression (LIC) methods~\cite{balle2016end,yang2023lossy,dupont2021coin, agustsson2023multi,choi2019variable,johnston2018improved} have introduced a paradigm shift by employing neural network-based joint optimization of transforms, quantization, and entropy modeling.
Ballé et al.~\cite{balle2016end} challenged the traditional paradigm by proposing an end-to-end trainable framework based on autoencoders, demonstrating the potential of jointly optimizing learned transforms and entropy estimation. 
Subsequent research has explored various avenues, including refinements in entropy estimation~\cite{balle2018variational, jiang2023mlic, minnen2020channel, duan2023qarv} and enhancements to transform representation~\cite{duan2023learned,koyuncu2024efficient,cuienhancing, zhang2024learnability, zhu2022transformer}.
LIC methods achieve superior rate-distortion (R-D) performance by learning information-rich latent representation and accurate entropy estimation, surpassing traditional codecs in numerous practical scenarios~\cite{kim2024c3, xu2025decouple, li2025callic, jeon2023context, zhang2025camsic, song2023efficient,liu2024slimmable,li2022region}.

\begin{figure*}[!t]
    \centering
    \subfloat[Standard Compression]{\includegraphics[trim={260pt 280pt 390pt 155pt},clip,width=0.47\textwidth]{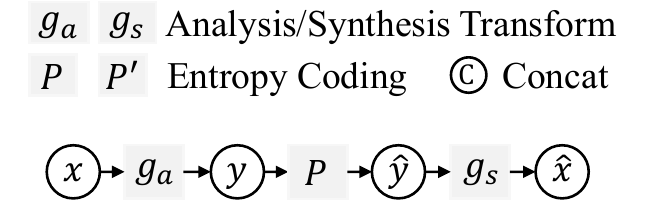}
    \label{fig:overview_a}}
    \hfill 
    \subfloat[Our proposed DCIC-sgp]{\includegraphics[trim={280pt 280pt 390pt 140pt},clip,width=0.45\textwidth]{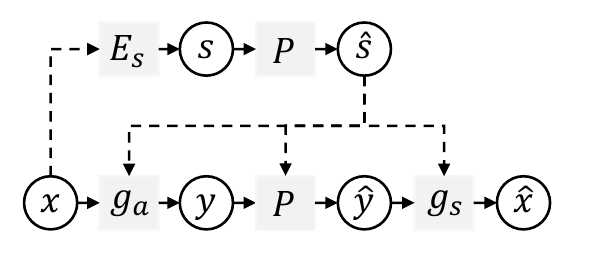}
    \label{fig:overview_b}}
    
    \caption{
    A conceptual comparison of compression paradigms. 
    (a) The standard framework, which relies on a single, entangled latent representation. 
    (b) Our proposed DCIC-sgp framework, which instantiates the principle of functional decomposition through a causally-dependent hierarchical architecture. In this paradigm, a Prior Extractor ($E_s$) generates a structure prior ($s$), which then holistically guides the entire pipeline by modulating the analysis transform ($g_a$), assisting the entropy model ($P$), and steering the synthesis transform ($g_s$).
    }
    \label{Figure:overview}
\end{figure*}

Despite significant progress, a foundational limitation pervades most LIC frameworks. 
The paradigm of directly encoding the image into a single latent representation, as illustrated in Figure~\ref{Figure:overview}(a), requires this monolithic representation to concurrently capture signal components with disparate statistical properties—the low-frequency, invariant object boundaries and the high-frequency, transient textures.
This forces a trade-off that is particularly detrimental at low bitrates. 
The consequences of this entanglement are particularly evident in the widely-adopted hyperprior framework~\cite{balle2018variational, duan2023qarv, hu2020coarse}, where the architectural choice of deriving side information from an already compacted latent representation gives rise to an inherent information bottleneck. This limitation in the available side information impairs the model's ability to intelligently preserve structural dependencies during the subsequent quantization step, thereby exacerbating the issue of geometric deformation.

These observations motivate the exploration of alternative paradigms. Instead of solely pursuing more powerful universal transforms~\cite{zou2022devil,lifrequency,liu2023learned} or more complex autoregressive entropy models~\cite{minnen2018joint, he2022elic, minnen2020channel,jiang2023mlic} to contend with the entangled information, a more efficacious path lies in functional decomposition: a principle of explicitly disentangling the stable, structural backbone of an image from its complex details. Such a methodology promises a more judicious allocation of bits and a more robust structural representation.

To this end, we propose Deeply-Conditioned Image Compression via self-generated priors (DCIC-sgp), a framework that instantiates this principle of functional decomposition. The core idea of our method is to first distill a potent, self-generated prior to encapsulate the image's structural backbone, which is then subsequently utilized not as mere side-information, but to holistically modulate the entire compression pipeline.
This approach is realized through a causally-dependent hierarchical architecture, as illustrated in Figure~\ref{Figure:overview}(b), and stands in stark contrast to multi-branch methods that rely on parallel, independent encoders~\cite{nakanishi2018neural, fu2024learned}.
This "deep conditioning" is deeply integrated to: (1) Modulate the analysis transform ($g_a$), liberating it to specialize in encoding the residual details; (2) Bolster the entropy model ($\mathcal{P}$) with rich, global context; and (3) Steer the synthesis transform ($g_s$) with a multi-scale fusion strategy. Among these, the conditioning of the analysis transform constitutes the most significant departure from prior art and is the key enabler of effective functional decomposition. This holistic approach differs from prior art where conditioning was often confined to later stages or based on information-limited priors~\cite{liu2025region,fu2024learned,nakanishi2018neural,ge2021hierarchical}.

The primary contributions of this paper are thus threefold:
\begin{itemize}

     \item We propose a new paradigm for learned image compression, predicated on the principle of \textit{functional decomposition}, to address the "information entanglement" that causes geometric deformation. This paradigm is realized through a novel, causally-dependent hierarchical architecture that explicitly decouples the representation of invariant structure from that of transient details.

    \item We present DCIC-sgp, a novel framework that instantiates the proposed paradigm through a \textit{holistic, end-to-end conditioning mechanism}. In this framework, a self-generated structural prior is systematically integrated to guide every critical stage of the pipeline: it modulates the analysis transform to enable specialization, provides global context to the entropy model, and steers the synthesis transform via multi-scale fusion.

    \item We provide extensive empirical validation demonstrating that our method not only substantially mitigates geometric artifacts—a key qualitative failure mode of existing codecs—but also establishes highly competitive rate-distortion performance across multiple standard benchmarks, achieving BD-rate savings of up to 15.7\% against VTM-12.1.
\end{itemize}

\section{Related Work}
\label{sec-related}
Modern learned image compression (LIC) has evolved significantly, with major advancements centered on improving three core components: the entropy model for accurate probability estimation, the transform for creating compact representations, and the overall architecture for effective information processing.
\subsection{Advances in Entropy Modeling}
Early LIC methods established the VAE-based framework with hyperpriors~\cite{balle2018variational}, which remains a cornerstone of the field. Subsequent work has primarily focused on capturing richer dependencies in the latent space to achieve more accurate entropy estimation. Autoregressive models, for example, have been extensively explored, evolving from computationally intensive spatially sequential approaches~\cite{minnen2018joint} to more efficient parallelized mechanisms~\cite{he2021checkerboard,he2022elic,fu2024fast} and channel-wise contexts~\cite{minnen2020channel,feng2023semantically, li2024groupedmixer}. Beyond autoregression, other powerful statistical models like Gaussian mixture models~\cite{cheng2020learned, zhang2025generalized} and multi-reference techniques~\cite{jiang2023mlic, jiang2025mlicv2} have been proposed. While these methods have become increasingly sophisticated, their effectiveness is ultimately bounded by the quality of the latent representation they are modeling.
Further innovations focus on enhancing the information available to the entropy model or optimizing the context itself. Lu et al.~\cite{lu2025learned} improve estimation by using cross-attention with an external, learnable dictionary derived from training data patterns. Han et al.~\cite{han2024causal} introduce a Causal Context Adjustment loss (CCA-loss) to explicitly guide the encoder to place more predictive information earlier in the autoregressive context. Our approach differs by using a self-generated internal prior for conditioning, rather than external dictionaries or context optimization losses.

\subsection{Advances in Transform Design}
The design of the analysis and synthesis transforms, which map images to and from the latent space, is another critical research frontier. To enhance representational capacity beyond simple CNNs, significant effort has been invested in designing advanced transform components. Attention modules~\cite{cheng2020learned, mishra2022deep, mital2023neural} and various Transformer-based architectures~\cite{zou2022devil, zhu2022transformer, liu2023learned} have been integrated to better capture long-range dependencies. Hybrid CNN-Transformer models~\cite{liu2023learned} and frequency-aware designs~\cite{lifrequency} further tailor the architecture to the specific characteristics of image data. 
Another direction integrates traditional transforms; Fu et al.~\cite{fu2024weconvene} embed Discrete Wavelet Transforms (DWT) within CNN layers (WeConv) and apply DWT before the entropy model (WeChARM) to explicitly handle frequency-domain redundancy. While also aiming to handle different signal characteristics, our method achieves functional decomposition through deep conditioning rather than explicit frequency-band separation.

Despite the sophistication of these single-path transform designs, their reliance on a single latent representation presents a core challenge.
The necessity of encoding all image information into one compact representation can create an information bottleneck, which may limit further gains in rate-distortion performance, especially for images with complex content. This motivates the investigation of alternative architectures that can overcome this intrinsic limitation.

\subsection{Conditional and Multi-Path Architectures}
To overcome the limitations of single latent representations, a separate line of research has explored conditional and multi-path architectures, which can be broadly categorized by the target and nature of their conditioning.

One common strategy focuses on the synthesis stage; for instance, Nakanishi et al.~\cite{nakanishi2018neural} fuse multi-scale features within the decoder to improve reconstruction quality. Another category targets the entropy model. The Hierarchical Compression model by Ge et al.~\cite{ge2021hierarchical} uses features from deeper layers to condition the probability estimation of shallower ones. The dual-branch encoder of Fu et al.~\cite{fu2024learned} exemplifies the parallel generation paradigm, where two independent encoders (e.g., using 3x3 and 1x1 convolutions) generate distinct representations simultaneously. While one stream provides side information for the other's entropy coding, their core feature extraction processes are decoupled. Other works involve sending explicit but information-sparse priors, such as the coarse segmentation masks used in segmentation-guided methods~\cite{liu2025region}.
Similarly, saliency maps, learned jointly with the compression task using multi-scale networks, have also been explored as auxiliary information to guide the compression process~\cite{mishra2021multi}.

While these strategies are varied, they largely exhibit what can be termed as "shallow conditioning." 
This manifests in two primary ways. Firstly, the conditional information is primarily leveraged to assist the later stages of the pipeline (i.e., entropy modeling or synthesis). 
Even when a method like Fu et al.~\cite{fu2024learned} uses a dual-branch representation to aid the entropy model, the integration is superficial: the two representations are simply concatenated along the channel axis, which is functionally equivalent to merely increasing the number of slices in a standard autoregressive model.
Secondly, the underlying representations are often generated through parallel and independent processes, precluding the possibility of one stream dynamically guiding the feature extraction of another.

Our work departs fundamentally from this paradigm. We introduce a hierarchical, causally-dependent architecture where the generation of the detail representation is explicitly conditioned on a pre-existing, rich structural prior. Even in cases like segmentation-guided methods where a prior influences the analysis transform, the prior itself is information-sparse and its integration is localized. In contrast, our framework is the first to establish a system where a potent conditional representation is first generated and then used to holistically guide the entire compression pipeline. This includes not only modulating the analysis transform towards a functional decomposition, but also steering the synthesis process through a sophisticated multi-scale fusion strategy. It is this "deep conditioning" paradigm—defined by the richness of the prior, the causal dependency of the representations, and the depth of its integration across both feature extraction and synthesis—that represents a distinct and more powerful architectural choice.

\subsection{Temporal Conditioning in Video Compression}
The principle of conditioning is also paramount in the adjacent field of learned video compression. State-of-the-art conditional codecs~\cite{li2021deep, li2023neural, li2024neural} leverage previously decoded frames as powerful temporal priors. In this paradigm, features from prior frames are used to directly modulate the encoding and decoding process of the current frame's features. This conditional context is deeply integrated to assist both the generation of the latent representation and its subsequent entropy modeling. While these methods naturally leverage a readily available temporal prior, our work addresses the more challenging task of single-image compression, where no such prior exists, by first creating a rich, spatial prior from the image itself to guide its own compression.

\section{Method}
\label{sec-method}
\subsection{Preliminaries: The Standard Compression Framework}
\label{sec:preliminaries}
Learned image compression is typically formulated within an end-to-end optimized framework~\cite{balle2016end}. 
As illustrated in Figure~\ref{fig:overview_a}, this standard paradigm consists of three primary components:
\begin{itemize}[noitemsep]
    \item An analysis transform, $g_a: \mathcal{X} \to \mathcal{Y}$, which maps the input image $x$ to a latent representation $y$.
    \item An entropy model, $\mathcal{P}$, which estimates the probability distribution of the quantized latents $\hat{y}$ to enable lossless coding.
    \item A synthesis transform, $g_s: \hat{\mathcal{Y}} \to \mathcal{X}$, which reconstructs the image $\hat{x}$ from the decoded latent representation $\hat{y}$.
\end{itemize}
The core data pipeline can be expressed as:
\begin{equation}
\label{eq:pipeline}
y = g_a(x), \quad \hat{y} = Q(y), \quad \hat{x} = g_s(\hat{y}),
\end{equation}
where $Q(\cdot)$ represents a quantization operation, such as rounding.

To accurately model the distribution of the latents, entropy models often employ a hyperprior architecture~\cite{balle2018variational, minnen2018joint}. A hyper-analysis transform $h_a$ generates a compact hyper-latent $z=h_a(y)$. The quantized hyper-latent $\hat{z}$ is then used by a hyper-synthesis transform $h_s$ to predict the parameters (e.g., mean $\mu_i$ and scale $\sigma_i$) of a probability distribution for each element of the latent representation. To model the discrete probabilities of the quantized values, this distribution, typically a Gaussian, is convolved with a unit uniform distribution:

\begin{equation}
\label{eq:entropy_model}
p(\hat{y} | \hat{z}) = \prod_{i} \left( \mathcal{N}(\hat{y}_i ; \mu_i(\hat{z}), \sigma_i^2(\hat{z})) \ast \mathcal{U}(-\tfrac{1}{2},\tfrac{1}{2}) \right).
\end{equation}

The entire framework is optimized end-to-end by minimizing the rate-distortion loss $\mathcal{L}$, which is the weighted sum of the bitrates $R$ for both the quantized latent and hyper-latent representations, and the distortion $D$:
\begin{equation}
\label{eq:standard_loss}
\mathcal{L} = R(\hat{y}) + R(\hat{z}) + \lambda D(x, \hat{x}),
\end{equation}
where $R(\cdot) = \mathbb{E}[-\log_2 p(\cdot)]$ and $\lambda$ is a hyperparameter controlling the trade-off. The probability of the hyper-latents, $p(\hat{z})$, is typically modeled with a simple factorized density estimation~\cite{balle2018variational}.

A key architectural limitation of this standard framework lies in its conditioning mechanism. The hyperprior $z$ is derived from the very latent representation $y$ it is meant to assist. This creates an inherent information bottleneck: the context available for probability estimation is fundamentally constrained by the information already encoded in the compact latent $y$. 
This motivates the exploration of a richer, more powerful source of conditioning, which is the cornerstone of our proposed framework.

\subsection{Overall Architecture of DCIC-sgp}
\label{sec:arch_overview}
\begin{figure*}[tbp]
\centering
\includegraphics[trim={35pt 110pt 85pt 90pt}, clip, width=\textwidth]{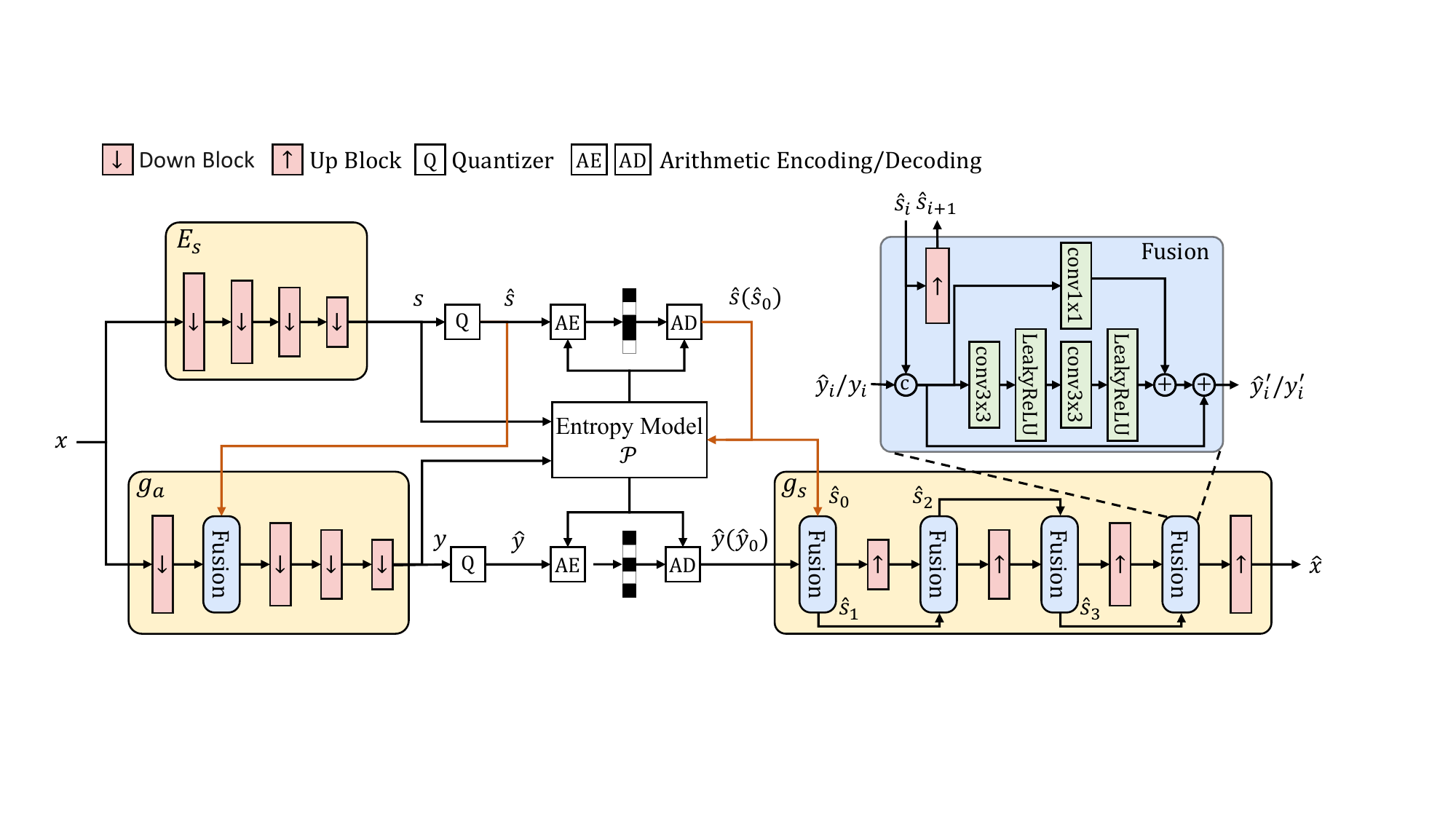}
\caption{
The overall framework of the proposed Deeply-Conditioned Image Compression with self-generated priors (DCIC-sgp) method. A Prior Extractor ($E_s$) maps the original image to a structure prior ($s$). The decoded prior ($\hat{s}$) is then used to guide both the Conditioned Analysis Transform ($g_a$) and the Entropy Model. The Conditioned Synthesis Transform ($g_s$) combines both representations ($\hat{s}$ and $\hat{y}$) for final reconstruction.
}
\label{fig:method_1}
\end{figure*}

To address the limitations of the standard framework, we introduce Deeply-Conditioned Image Compression with self-generated priors (DCIC-sgp), a framework architected to instantiate the principle of functional decomposition through a causally-dependent hierarchical process, as illustrated in Figure~\ref{fig:method_1}.

The central tenet of DCIC-sgp is to first leverage a Structure Information Extractor ($E_s$) to distill a potent, self-generated prior ($s$) that encapsulates the image's structural backbone. This powerful prior is then utilized to holistically guide the subsequent compression pipeline. Specifically, the decoded prior ($\hat{s}$) dynamically modulates the Conditioned Analysis Transform ($g_a$) to generate a detail-oriented representation ($y$). Both representations ($s$ and $y$) are then efficiently compressed by a unified Entropy Model ($\mathcal{P}$). Finally, during reconstruction, the Conditioned Synthesis Transform ($g_s$) leverages both decoded representations ($\hat{s}$ and $\hat{y}$) in a multi-scale fusion strategy to produce the final output.

The key novelty of DCIC-sgp thus lies not merely in using a prior, but in establishing this explicit, hierarchical dependency where the generation of the detail representation $y$ is fundamentally conditioned on the structure prior $s$.

\subsection{Structure Information Extractor}
\label{sec:prior_extractor}
The initial stage of the DCIC-sgp framework is the Prior Extractor ($E_s$), a dedicated network designed to distill a rich structural prior, $s$, from the input image $x$. This network is specifically architected to capture the image's invariant global information, such as object contours and overall layout, which forms the basis for our functional decomposition. 

The resulting representation $s$ is then quantized and entropy-coded into a bitstream using a standard hyperprior-based method, the specifics of which will be detailed in Section~\ref{sec:entropy_model}. Finally, the decoded structure information, $\hat{s}$, is made available to all subsequent components of the compression pipeline, serving as the powerful guiding prior for the encoding of detail-oriented information.

A pertinent question is how this functional decomposition is learned without an explicit supervisory signal for structure. The decomposition is not explicitly enforced, but rather emerges from the causal dependency architected into our framework, wherein the generation of the detail representation $y$ is conditioned on the prior $s$. This dependency compels the interconnected networks, $E_s$ and $g_a$, to learn a collaborative division of labor during end-to-end training. The training process naturally guides a co-evolution of the two representations, where $E_s$ is encouraged to encapsulate the most compressible, low-entropy information (i.e., structure), thereby liberating $g_a$ to efficiently encode the more complex residual details.

\subsection{Conditioned Analysis and Synthesis Transforms}
\label{sec:conditioned_transforms}
While the Prior Extractor captures the image's structural backbone, the subsequent transforms ($g_a$, $g_s$) are responsible for efficiently encoding and decoding the remaining textural and detailed information. Their operations are deeply conditioned by the decoded structure prior $\hat{s}$ to ensure both compression efficiency and high reconstruction fidelity.
\subsubsection{Conditioned Analysis Transform}
\label{sec:cond_analysis}
In contrast to parallel approaches with independent encoders, our causally-dependent framework employs an analysis transform, $g_a$, whose feature extraction process is dynamically modulated by the structure prior $\hat{s}$.
Specifically, the prior $\hat{s}$ is spatially upsampled to match the dimensions of a target feature map within the initial layers of $g_a$. The upsampled prior and the target feature map are then jointly fed into a \textit{Fusion Module}, which first concatenates them along the channel axis and subsequently processes the combined tensor through several convolutional layers to seamlessly integrate the structural information. 
With the prior $\hat{s}$ providing the global, low-frequency structural information, this mechanism allows the subsequent, deeper layers of $g_a$ to focus their capacity on efficiently capturing the remaining high-frequency information, such as complex local textures and fine details.
This functional decomposition is the key to mitigating geometric deformation, as the model avoids representing both stable structure and volatile details within a single, heavily compressed latent variable. This process yields the latent detail representation $y = g_a(x, \hat{s})$.
\subsubsection{Conditioned Synthesis Transform}
\label{sec:cond_synthesis}
The structure prior $\hat{s}$ plays a complementary and crucial role during reconstruction.
It is hierarchically fused with the decoded detail representation $\hat{y}$ at multiple scales within the synthesis transform $g_s$. 
This ensures that the robust global structure from $\hat{s}$ anchors the reconstruction process, preventing the kind of large-scale geometric deformation that can occur when decoding from a single, less-structured latent representation. 
This strategy actively guides the rendering of fine details and ensures the final reconstruction, $\hat{x} = g_s(\hat{y}, \hat{s})$, maintains high structural and textural fidelity.

\subsection{Entropy Model}
\label{sec:entropy_model}
\begin{figure}[tbp]
\centering
\includegraphics[trim={100pt 190pt 100pt 168pt}, clip, width=\textwidth]{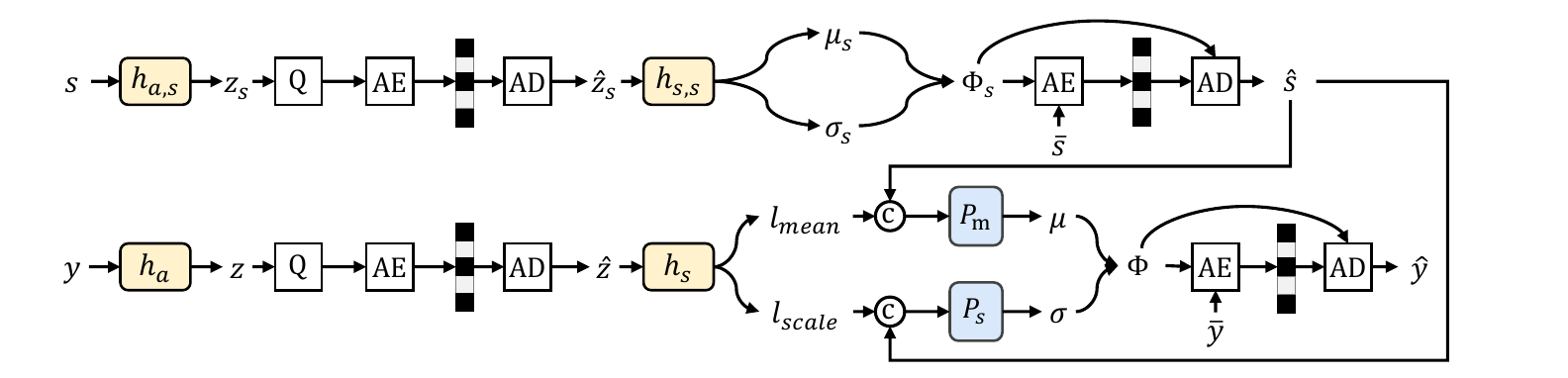}
\caption{
Our unified Entropy Model. It features two distinct processing paths: one for the structure prior $s$ (top), which employs a standard hyperprior mechanism, 
and another for the detail representation $y$ (bottom), which uses a conditional mechanism. 
To be precise with the notation used in the figure, $\bar{s}$ and $\bar{y}$ represent the quantized latents before entropy coding, while $\hat{s}$ and $\hat{y}$ represent the latents after entropy decoding.
Due to lossless entropy coding, they are numerically identical (e.g., $\bar{s} = \hat{s}$). 
The parameter networks $P_m$ and $P_s$ fuse information from the hyper-decoder with the decoded structure prior $\hat{s}$ to generate the final distribution parameters for $y$.
}
\label{fig:method_2}
\end{figure}

The entropy model in DCIC-sgp, denoted as $\mathcal{P}$, is responsible for the efficient entropy coding of the two distinct latent representations: the structure prior $s$ and the detail representation $y$. 
To handle each effectively according to its role, the model adapts its conditioning strategy. It employs a standard hyperprior mechanism for the self-contained structure prior, while a novel conditional approach is used for the detail representation, leveraging two complementary sources of information. The two mechanisms are illustrated in Figure~\ref{fig:method_2} and detailed below.

\subsubsection{Entropy Modeling for structure prior}
The structure prior $s$ is compressed using a standard hyperprior model, identical to the one described in Section~\ref{sec:preliminaries}. 
A hyper-analysis transform, $h_{a,s}$, maps the structure prior $s$ to its hyper-latent $z_s$. 
The quantized hyper-latent $\hat{z}_s$ is then used by a hyper-synthesis transform, $h_{s,s}$, to predict the parameters $\Phi_s = (\mu_s, \sigma_s^2)$ of a Gaussian distribution that models the probability of the quantized representation $\hat{s}$.
This process is formalized by the following equation:
\begin{equation}
    \hat{z}_s = Q(h_{a,s}(s)), \quad \Phi_s = (\mu_s, \sigma_s^2) = h_{s,s}(\hat{z}_s).
\end{equation}
\subsubsection{Conditional Entropy Modeling for Detail Representation}
\label{sec:entropy_y}
The core of our entropy modeling lies in the probability estimation for the detail representation $y$.
To achieve the most accurate estimation, we condition the model on two complementary sources of information: its own compact hyper-latent $\hat{z}_y$ and the rich, decoded structure prior $\hat{s}$. 

The rationale behind this dual-source conditioning is to synergize two complementary information streams. The hyper-prior $\hat{z}_y$, derived from $y$, provides highly correlated, localized statistics specific to the textural details encoded within y. In parallel, the structure prior $\hat{s}$ offers a rich, global context that captures the underlying structural dependencies of the image, information which is unavailable to $\hat{z}_y$.

As depicted in the bottom part of Figure~\ref{fig:method_2}, the final distribution parameters for $y$, $\Phi_y = (\mu_y, \sigma_y)$, are generated by dedicated parameter networks ($P_m, P_s$) that take both the intermediate features from its hyper-synthesis transform, $h_{s,y}$, and the structure prior $\hat{s}$ as input:
\begin{equation}
\begin{aligned}
    \hat{z}_y = Q(h_{a,y}(&y)),\quad l_{\text{mean}}, l_{\text{scale}} = h_{s,y}(\hat{z}_y), \\
    \Phi_y = (\mu_y, \sigma_y) &= (P_m(l_{\text{mean}}, \hat{s}), P_s(l_{\text{scale}}, \hat{s})).
\end{aligned}
\end{equation}
By leveraging both local and global contexts, this conditional model achieves a more robust and accurate probability estimation.
\subsection{Optimization Objective}
\label{sec:optimization}
The entire DCIC-sgp framework, including the Prior Extractor ($E_s$), the Conditioned Transforms ($g_a, g_s$), and the Entropy Model ($\mathcal{P}$), is trained end-to-end by minimizing a single rate-distortion loss function, $\mathcal{L}$. This objective function sums the bitrates required to encode both the structure and detail representations with their respective hyper-latents, along with a term for the distortion between the original and reconstructed images.
Formally, the total loss $\mathcal{L}$ is defined as:
\begin{equation}
\label{eq:final_loss}
\mathcal{L} = \underbrace{R(\hat{s}) + R(\hat{z}_s)}_{\text{Structure Rate}} + \underbrace{R(\hat{y}) + R(\hat{z}_y)}_{\text{Detail Rate}} + \lambda D(x, \hat{x}),
\end{equation}
where $R(\cdot) = \mathbb{E}[-\log_2 p(\cdot)]$ represents the bitrate estimated from the probability distributions $p(\hat{s}|\hat{z}_s)$ and $p(\hat{y}|\hat{s}, \hat{z}_y)$. The term $D(x, \hat{x})$ is the distortion measure, such as MSE or MS-SSIM. The Lagrange multiplier $\lambda$ controls the trade-off between the total rate and the distortion. 

\section{Experimental result}
\subsection{Experimental Setup}
\label{sec:exp_setup}
\subsubsection{Datasets}
\label{sec:datasets}
All models were trained on a large-scale dataset of approximately 300,000 images sampled from the ImageNet-1K dataset~\cite{deng2009imagenet}. During training, images were randomly cropped to $256 \times 256$ patches. For evaluation, we used three standard benchmark datasets: the Kodak dataset~\cite{KodakDataset} (24 images, 768$\times$512), the CLIC Professional Validation dataset~\cite{clic2021} (41 images, up to 2K resolution), and the Tecnick dataset~\cite{asuni2014testimages} (100 images, 1200$\times$1200).

\subsubsection{Implementation Details}
\label{sec:implementation}
To demonstrate the effectiveness and versatility of our DCIC-sgp paradigm, we implemented two primary versions of our framework. Each version is designed for direct comparison against a key baseline representing a different performance tier: the foundational Mean \& Scale Hyperprior (MSH)~\cite{minnen2018joint} and the high-performance TCM~\cite{liu2023learned}.

A core principle of our experimental design is fairness; therefore, each DCIC-sgp variant strictly follows the training protocol of its corresponding baseline. 
For the DCIC-sgp-MSH model, we adopted the training configuration from the CompressAI library's MSH implementation, including its optimizer settings, learning rate schedule, and Lagrange multipliers ($\lambda$). 

Similarly, for the DCIC-sgp-TCM model, we followed the training strategy detailed in the original TCM publication~\cite{liu2023learned}, including their specific learning rate decay schedule and $\lambda$ values. 
To generate the R-D curves, our models were trained for two different distortion metrics, MSE and MS-SSIM, corresponding to the distortion term $D$ in Eq.~\eqref{eq:final_loss}. 
For the MSE-optimized models,the DCIC-sgp-MSH used $\lambda \in \{0.003,0.005,0.01,0.025,0.05\}$, and the DCIC-sgp-TCM used $\lambda \in \{0.0035,0.0067,0.013,0.025,0.05\}$. 
For the MS-SSIM-optimized models, the DCIC-sgp-MSH used $\lambda \in \{3,5,8,16,36,64\}$, and the DCIC-sgp-TCM used $\lambda \in \{2,5,10,25,50\}$.
Crucially, to maintain this principle of fairness and rigorously isolate the performance gains of our proposed paradigm, the architecture of our Prior Extractor ($E_s$) is designed to be identical to the analysis transform ($g_a$) of the corresponding baseline, which typically consists of four down blocks.
This design choice ensures that our performance improvements are attributable to the proposed functional decomposition and deep-conditioning framework itself, rather than to the use of a different, potentially more powerful, network backbone.
For architectural transparency, the specific configurations for $E_s$ corresponding to each baseline are detailed in Table~\ref{tab:es_arch}.
Correspondingly, the Up Blocks employ approximate inverse operations.
In all our DCIC-sgp implementations, the channel count for both the structure prior ($s$) and the detail representation ($y$) was set to 320, while their respective hyper-latents ($\hat{z}_s$ and $\hat{z}_y$) were set to 192.

\begin{table}[t]
\centering
\caption{
The $E_s$ architecture is set to be identical to the analysis transform ($g_a$) of the corresponding baseline (MSH~\cite{minnen2018joint} or TCM~\cite{liu2023learned}), which typically consists of four down blocks, for a fair comparison.
Abbreviations: RBS (ResidualBlockWithStride), TCM block (Transformer-CNN Mixture block, as defined in ~\cite{liu2023learned}).
}

\begin{tabular}{@{} l @{\hspace{1.5cm}} l @{}} 
    \hline
    \textbf{DCIC-sgp-MSH}  & \textbf{DCIC-sgp-TCM} \\
    \hline
    
    \makecell[tl]{
    Conv: 5x5 c192 s2, GDN \\
    Conv: 5x5 c192 s2, GDN \\
    Conv: 5x5 c192 s2, GDN \\
    Conv: 5x5 c320 s2
    }
    
    & 
    \makecell[tl]{
    RBS: 3x3 c256 s2, TCM block \\
    RBS: 3x3 c256 s2, TCM block \\
    RBS: 3x3 c256 s2, TCM block \\
    Conv: 3x3 c320 s2
    }
    \\
    \hline
    \end{tabular}
    \label{tab:es_arch}
\end{table}

To ensure a robust and transparent comparison, the results for the baseline models were established as follows. 
The performance of the MSH baseline was reproduced by us using the public CompressAI library, ensuring an identical training and evaluation environment for our DCIC-sgp-MSH model. 
For the TCM baseline, to ensure the most accurate point of comparison, we report the official results published by its authors for their highest-performing public model, which corresponds to their configuration with 128 output channels (N=128), as obtained from their official codebase.
For all other state-of-the-art methods shown in the R-D curves, the performance data was obtained from their respective original papers or officially released codebases.

\subsubsection{Evaluation Metrics}
\label{sec:metrics}
We evaluated the performance of all methods using established, standard metrics. The rate was measured in bits per pixel (bpp). The distortion was measured using the Peak Signal-to-Noise Ratio (PSNR) and the Multi-Scale Structural Similarity Index (MS-SSIM), where higher values indicate better quality. For a comprehensive comparison of rate-distortion efficiency, we also report the Bjontegaard Delta Rate (BD-rate)~\cite{bjontegaard2001calculation} savings relative to the VTM-12.1 anchor.

\begin{figure*}[!t] 
    \centering
    \subfloat[PSNR on Kodak dataset]{\includegraphics[width=0.485\textwidth]{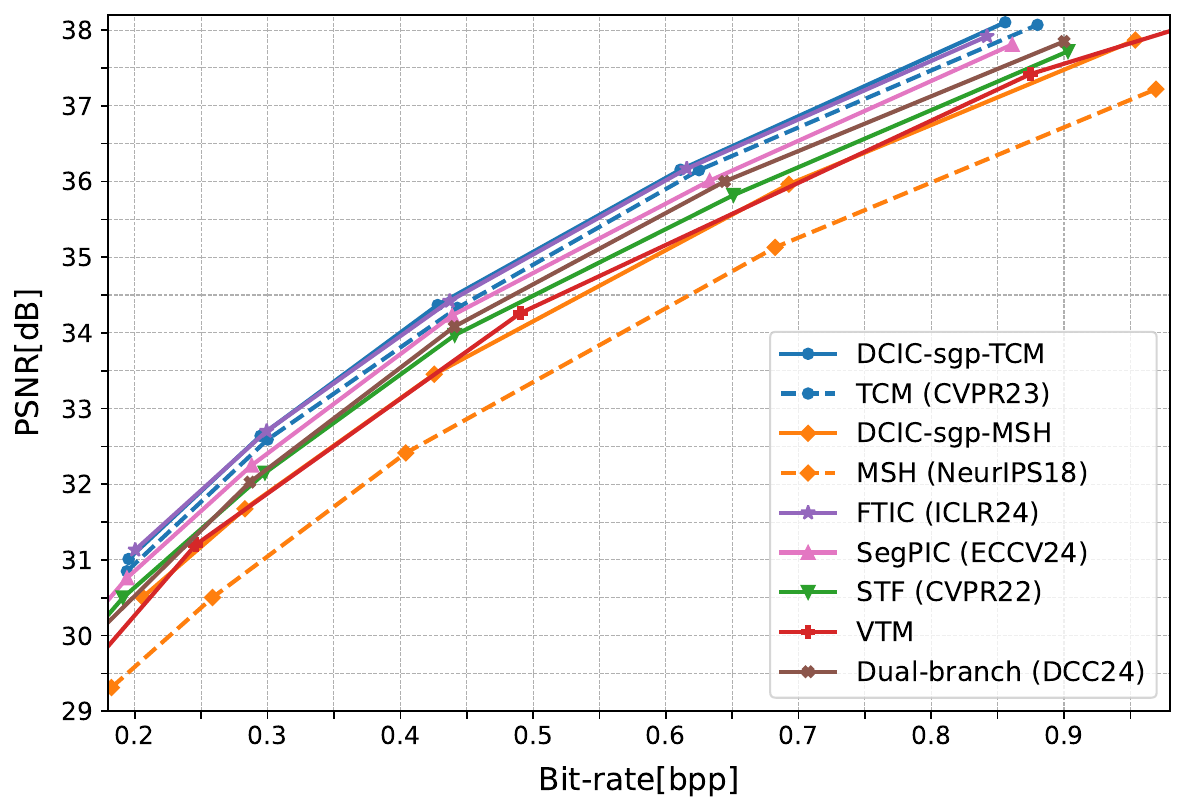}
    \label{fig:kodak_psnr}}
    \hfill 
    \subfloat[PSNR on CLIC-Pro-Valid dataset.]{\includegraphics[width=0.485\textwidth]{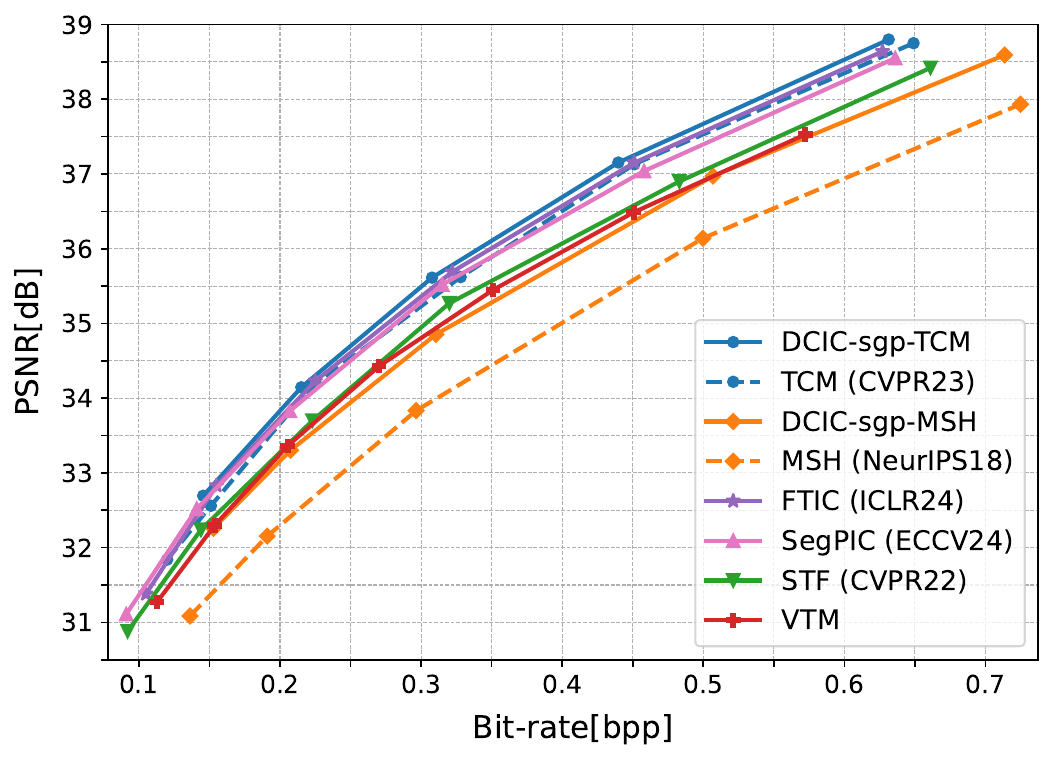}
    \label{fig:clic_psnr}}

    \subfloat[PSNR on Tecnick dataset]{\includegraphics[width=0.485\textwidth]{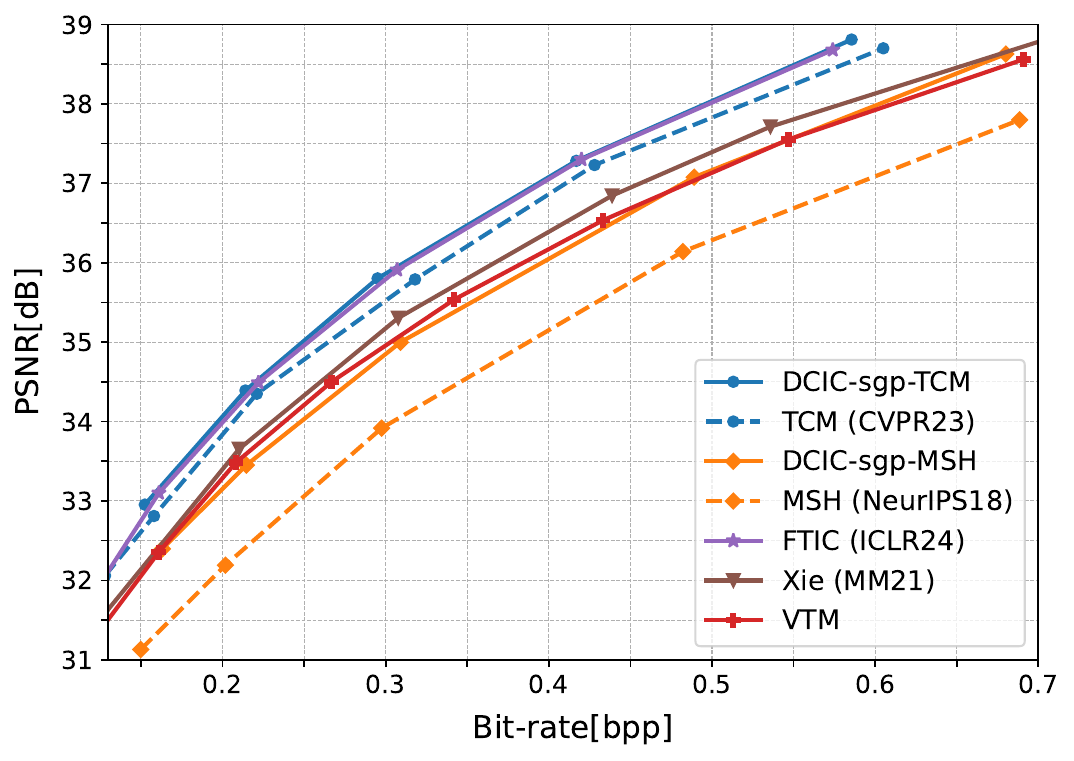}
    \label{fig:tecnick_psnr}}
    \hfill 
    \subfloat[MS-SSIM on Kodak dataset]{\includegraphics[width=0.485\textwidth]{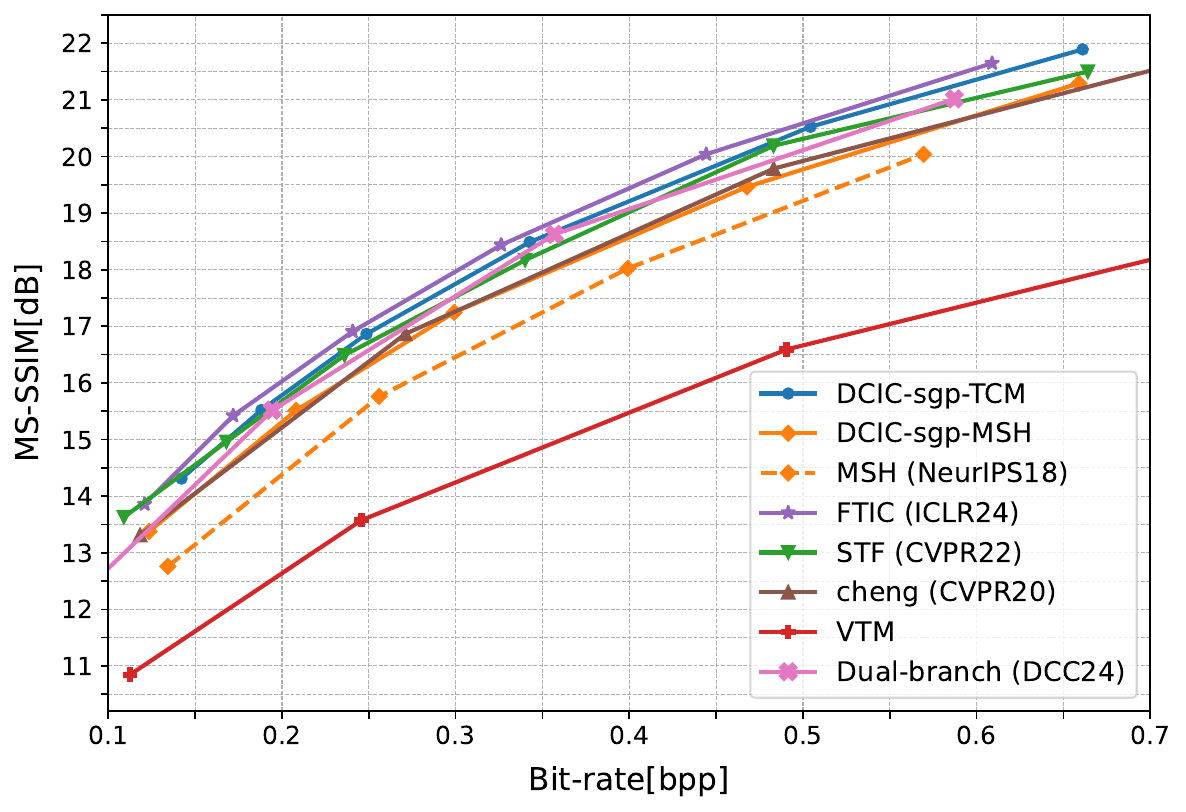}
    \label{fig:kodak_ms-ssim}}
    
    \caption{Rate-distortion performance of our DCIC-sgp models (DCIC-sgp-MSH and DCIC-sgp-TCM) compared against their respective baselines and other leading methods across various datasets and metrics.
    }
    \label{fig:main_performance_comparison}
\end{figure*}
\subsection{Rate-Distortion Performance}
\label{sec:rd_performance}
The rate-distortion (R-D) performance of our DCIC-sgp framework, compared against baselines and other leading methods, is presented in Figure~\ref{fig:main_performance_comparison}.
The results demonstrate the broad effectiveness of the deep conditioning paradigm. When our DCIC-sgp architecture is built upon a simple MSH-style backbone (denoted as DCIC-sgp-MSH), it yields a significant PSNR improvement of approximately 0.8dB over the original MSH. 
More impressively, when implemented with a high-performance TCM-style backbone (denoted as DCIC-sgp-TCM), 
our approach provides an additional gain of approximately 0.2dB, achieving highly competitive performance, especially on the challenging CLIC dataset. 
In terms of overall efficiency, our method achieves substantial BD-rate reductions of 14.4\%, 15.7\%, and 15.1\% compared to VTM on the Kodak, CLIC, and Tecnick datasets, respectively.

\begin{figure}[!t]
\centering
\includegraphics[trim={0pt 0pt 0pt 0pt}, clip, width=\textwidth]{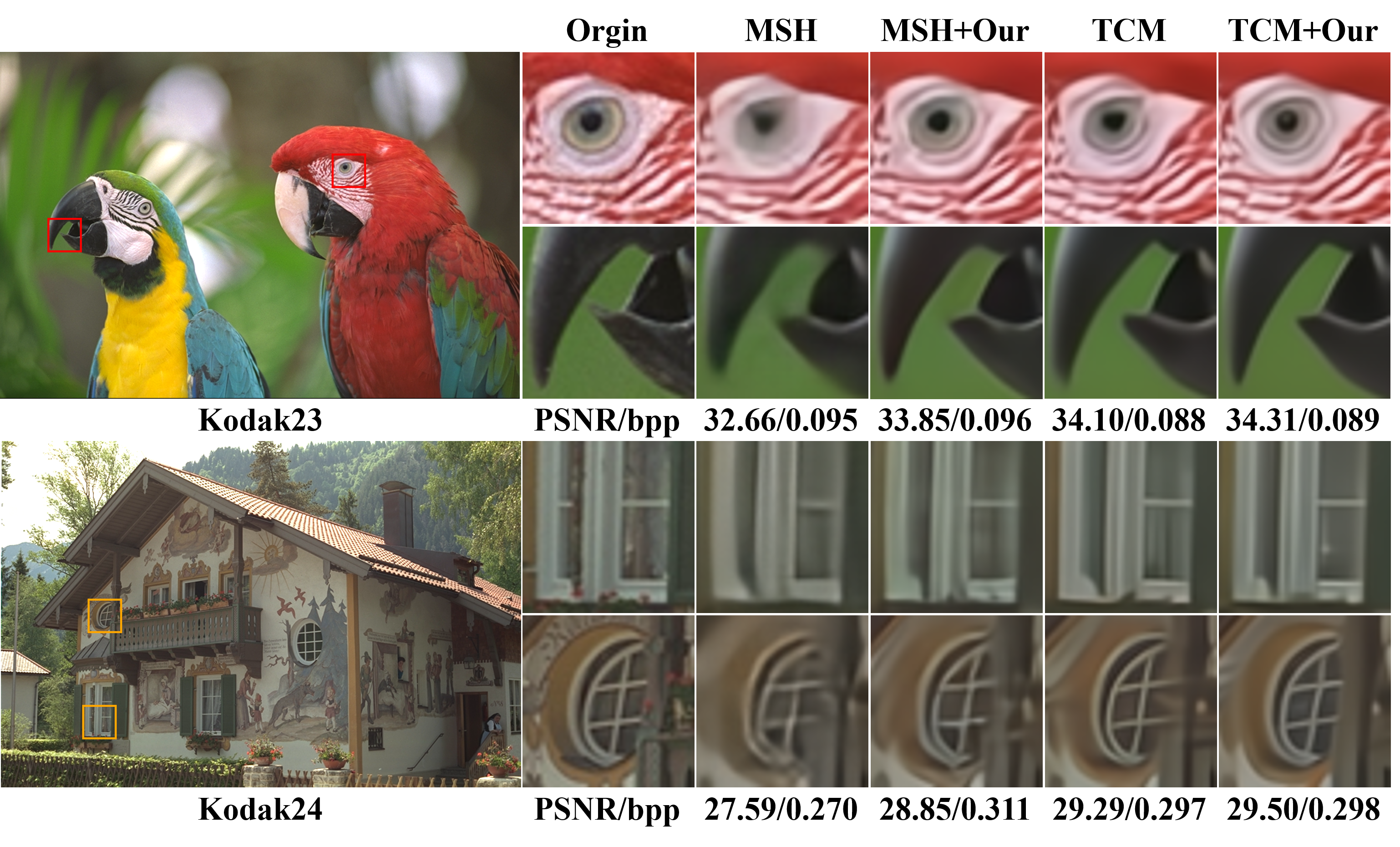}
\caption{
Visual comparison of reconstructed images by our method against baseline methods.
The close-ups highlight the superior ability of our DCIC-sgp framework to mitigate geometric deformation.
}
\label{fig:vis}
\end{figure}

Beyond objective metrics, Figure~\ref{fig:vis} provides compelling visual validation of our approach's primary advantage in mitigating geometric deformation. Notably, even when our DCIC-sgp-MSH model operates at a lower PSNR than the powerful TCM baseline (e.g., 33.85 dB vs. 34.10 dB on Kodak23), its reconstruction exhibits significantly less geometric deformation. 
As seen in the close-ups, the structural integrity of the parrot's eye and beak is much better preserved by our method.
This superior preservation of object integrity can be attributed to our core design principle: by first encoding and then leveraging a robust structural prior ($s$), our framework maintains object integrity even at very low bitrates—a critical challenge in learned image compression.

\subsection{Ablation Studies and Component Analysis}
\label{sec:ablation}
To dissect the contributions of the key components within our DCIC-sgp framework, we conducted a series of ablation studies. 
These experiments were performed on our DCIC-sgp-MSH model, allowing for a controlled comparison against the MSH baseline, which has a 0\% BD-rate, on the Kodak dataset.

\begin{table}[t]
 \centering
 \caption{
 Ablation study on the core components of DCIC-sgp-MSH, evaluated on the Kodak dataset. All runtimes were benchmarked on an NVIDIA RTX 3090 GPU.
 A lower BD-rate indicates better RD-performance. "$\mathcal{P}$" refers to Entropy Model.
 Total time is the sum of encoding and decoding time.
 }
 \begin{tabular}{@{} l c c c @{}} 
    \hline
    Methods & Total Time (ms) & \#Params (M) & BD-rate \\
    \hline
    DCIC-sgp (Full Model) &  258 & 55.85 & -12.09\% \\
    w/o Conditional $g_a$ &  241 &  54.74 & -7.86\% \\
    w/o Conditional $g_s$ &  238 &  49.45 & -9.17\% \\
    w/o Conditional Transforms &  225 & 48.35 & -6.22\% \\
    w/o Structure Prior ($\hat{s}$) in $\mathcal{P}$  & 257 & 42.61 & -8.29\% \\
    w/o Hyperprior ($\hat{z}_y$) in $\mathcal{P}$ & 259 & 45.31 & -11.10\% \\
    \hline
    \end{tabular}
    \label{tab:ablation_components}
\end{table}
\subsubsection{Analysis of Core DCIC-sgp Components}
Table~\ref{tab:ablation_components} indicates that the deep conditioning of the transforms is the most critical contributor to the model's performance, particularly the conditioning of the analysis transform ($g_a$).
Specifically, when we remove the conditional injection of the prior $\hat{s}$ from the analysis transform alone, denoted as w/o Conditional $g_a$, the BD-rate gain reduces from -12.09\% to -7.86\%. Removing it from the synthesis transform alone, denoted as w/o Conditional $g_s$, reduces the gain to -9.17\%. When conditioning is removed from both the analysis and synthesis transforms, denoted as w/o Conditional Transforms, the gain plummets further to -6.22\%.
These results indicate that the deep conditioning of the transforms is the most critical contributor to the model's performance, particularly the conditioning of the analysis transform ($g_a$).

Furthermore, the analysis of the entropy model components demonstrates that the two conditioning signals are valuable and complementary. Removing the rich structure prior $\hat{s}$ reduces the gain to -8.29\%, while removing the standard hyper-prior $\hat{z}_y$ only reduces it to -11.10\%, indicating that the global context from $\hat{s}$ provides a more significant contribution. 

Crucially, this component-wise analysis demonstrates that the overall performance gain is achieved through the targeted contributions of each distinct mechanism, each justifying its modest parameter cost with a substantial improvement in R-D efficiency. The overall parameter efficiency of the DCIC-sgp paradigm itself is further analyzed in the next section.

\subsubsection{Ablation Study on Prior Acquisition}
\label{sec:ablation_prior_acquisition}
To analyze the importance of the input signal's informational content for prior extraction, we investigated the plausible hypothesis that using a downsampled, low-resolution image as the input to the Prior Extractor ($E_s$) might be a more direct strategy for capturing global structures.To test this, we conducted two experiments against our DCIC-sgp-MSH baseline.

In the first experiment, the Prior Extractor ($E_s$) received a low-resolution version of the image, downsampled by a factor of 2, as its direct input, instead of the original full-resolution image. This setup resulted in a significant BD-rate degradation of 3.13\%.
In the second experiment, we explored if this low-resolution information could serve as a helpful auxiliary signal to the prior generation process itself. To test this, we augmented the Prior Extractor ($E_s$) so that it received both the original full-resolution image $x$ and its downsampled version as inputs, with the goal of distilling a single, potentially improved structural prior $s$. Ihis approach also harmed performance, leading to a BD-rate degradation of 1.87\%.

These results offer a crucial insight into how our framework achieves functional decomposition by decoupling information from the source. The first experiment demonstrates that providing the Prior Extractor ($E_s$) with a pre-filtered, low-resolution input hinders, rather than helps, its ability to effectively decouple the underlying structural information. Furthermore, the second experiment shows that attempting to assist the main compression stream with an additional, crudely-decoupled prior also degrades performance, suggesting this auxiliary signal acts as a source of interference.
Taken together, these findings validate our core principle: the most effective approach is to provide the network with the complete, original signal and empower it to learn the optimal strategy for disentangling the structural prior from the textural details itself, rather than relying on manual pre-filtering or information-degraded inputs.

\subsubsection{Architectural Efficiency}
To demonstrate that our performance gains stem from architectural intelligence rather than parameter scaling, we compare our DCIC-sgp-MSH model against the MSH baseline augmented with a channel-wise autoregressive entropy model (MSH+CH). To ensure a fair comparison of representational capacity, we set the channel count of the transform in MSH+CH to be the sum of the channel counts of our structure ($s$) and detail ($y$) representations.

The results are shown in Table~\ref{tab:ab_ch}. Our DCIC-sgp model achieves a -12.09\% BD-rate reduction with only 55.85M parameters, far surpassing the MSH+CH model which only achieves -6.70\% with a much larger 83.40M parameters. 
This clearly indicates the superior parameter efficiency of our functional decomposition approach compared to the brute-force strategy of simultaneously increasing channel complexity of the transform and employing an autoregressive entropy model.

\begin{table}[tbp]
\centering
\caption{
Comparison of architectural efficiency and the effect of iterative refinement for our DCIC-sgp-MSH model on the Kodak dataset. 
MSH+CH denotes the MSH baseline augmented with both increased transform channel capacity and a channel-wise autoregressive model, representing a brute-force complexity increase.
}
\label{tab:ab_ch}
\small 
\begin{tabular}{@{} l c c c @{}} 
    \toprule
    Method & Time (ms) & \#Params (M) & BD-rate (\%) \\
    \midrule
    \makecell[l]{MSH+CH} & 259 & 83.40 & -6.70\% \\
    \midrule
    DCIC-sgp-MSH (1 Iteration) & 258 & 55.85 & -12.09\% \\
    DCIC-sgp-MSH (2 Iterations) & 406 & 94.13 & -12.40\% \\
    DCIC-sgp-MSH (3 Iterations) & 570 & 132.41 & -12.61\% \\
    \bottomrule
\end{tabular}
\end{table}

\subsubsection{Analysis of Iterative Application}
Our framework's design naturally allows for iterative application, where the output of one pass can serve as an enhanced prior for a subsequent pass. We explored this characteristic, and the results are presented in Table~\ref{tab:ab_ch}. While applying the process for two or three iterations yields minor, incremental gains in BD-rate, the improvements come at a significant cost in terms of parameters and computational time. This observation suggests that the primary performance benefit of our framework is captured within the initial, single-pass application, further highlighting the efficiency of our core architectural design.

\begin{table}[tbp]
\centering
\caption{
Complexity and performance comparison against leading methods. Runtimes are reported in seconds per image (s/im) and BD-rate savings (\%) are evaluated against VTM-12.1 on the Kodak and CLIC datasets. All models were evaluated on an NVIDIA RTX 3090 GPU. Our "DCIC-sgp" entry corresponds to the highest-performing DCIC-sgp-TCM model.
}
\label{tab:sota_complexity}
\footnotesize
\setlength{\tabcolsep}{3pt}
\begin{tabular*}{\linewidth}{@{\extracolsep{\fill}} l c c c c @{}}
\toprule
Method & Enc. Time (s) & Dec. Time (s) & \#Params (M) & BD-rate (\%)~($\downarrow$) \\
\midrule
TCM~\cite{liu2023learned}      & 0.293 / 1.639 & 0.278 / 1.525 & 76.6 & -10.35 / -14.52 \\
FTIC~\cite{lifrequency}     & 249 / 937     & 247 / 910     & 71.0 & -10.94 / -12.68 \\
SegPIC~\cite{liu2025region}   & 0.137 / 0.593 & 0.149 / 0.472 & 83.5 & -7.03 / -10.86  \\
\midrule
DCIC-sgp (Ours) & 0.721 / 3.313 & 0.901 / 3.969 & 174.2 & -12.89 / -16.81\\
\bottomrule
\end{tabular*}
\end{table}
\subsection{Performance and Complexity Analysis} 
\label{sec:sota_comparison}
To provide a comprehensive evaluation, we compare our DCIC-sgp-TCM, against recent leading methods not only on rate-distortion performance but also on model complexity. To ensure a fair and rigorous comparison, all competing methods were evaluated under a unified setting. Specifically, we used the official, open-sourced model weights provided by the original authors for all methods. 

The highest-performance version of the TCM model, a key point of comparison, only had one of its highest rate points publicly available. Therefore, we used this specific TCM model as the anchor point. For all other methods (e.g., FTIC~\cite{lifrequency}, SegPIC~\cite{liu2025region}), we selected their officially released model weights corresponding to the rate point closest to this TCM anchor. 
This ensures that the comparisons presented in Table~\ref{tab:sota_complexity} are as fair and direct as possible.We report model parameters, encoding/decoding times measured on an NVIDIA RTX 3090 GPU, and BD-rate savings calculated against the VTM-12.1 anchor on both the Kodak and CLIC datasets.

The results in Table~\ref{tab:sota_complexity} reveal important insights into the practical trade-offs of different compression paradigms. 
For instance, the FTIC method~\cite{lifrequency}, which relies on an element-wise autoregressive model, promises strong theoretical performance.
However, an evaluation of the official implementation under a full end-to-end timing protocol highlights a critical challenge:the sequential nature of their entropy coding results in encoding and decoding times that are orders of magnitude slower, rendering them impractical for many applications.
Furthermore, under this rigorous evaluation, its rate-distortion performance did not surpass other leading methods like TCM. 
This highlights the importance of evaluating both theoretical potential and the runtime performance of a full, end-to-end practical implementation.
In contrast, our DCIC-sgp-TCM model achieves superior BD-rate savings while operating at a much more practical speed. 
The justification for this runtime performance lies in its architectural efficiency, not in a brute-force autoregressive extension. 

\begin{figure}[tbp]
    \centering
    \includegraphics[trim={0pt 240pt 0pt 60pt}, clip, width=\textwidth]{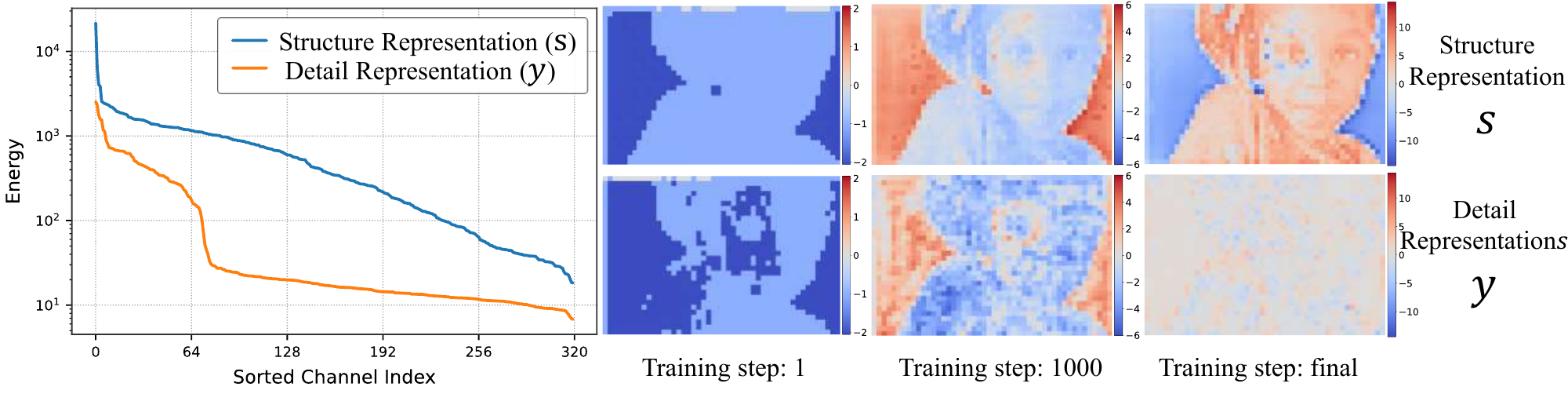}
    \caption{
    Visualization of the learned latent representations. Left: Energy distribution across channels for the structure prior $s$ and the detail representation $y$. Right: Visualization of the channels with maximum energy at different training stages, illustrating the co-evolution of the Prior Extractor and the Conditioned Analysis Transform.
    }
    \label{fig:vis_feature}
\end{figure}
\begin{figure}[!th]
    \centering
    \includegraphics[trim={0pt 0pt 0pt 0pt}, clip, width=\textwidth]{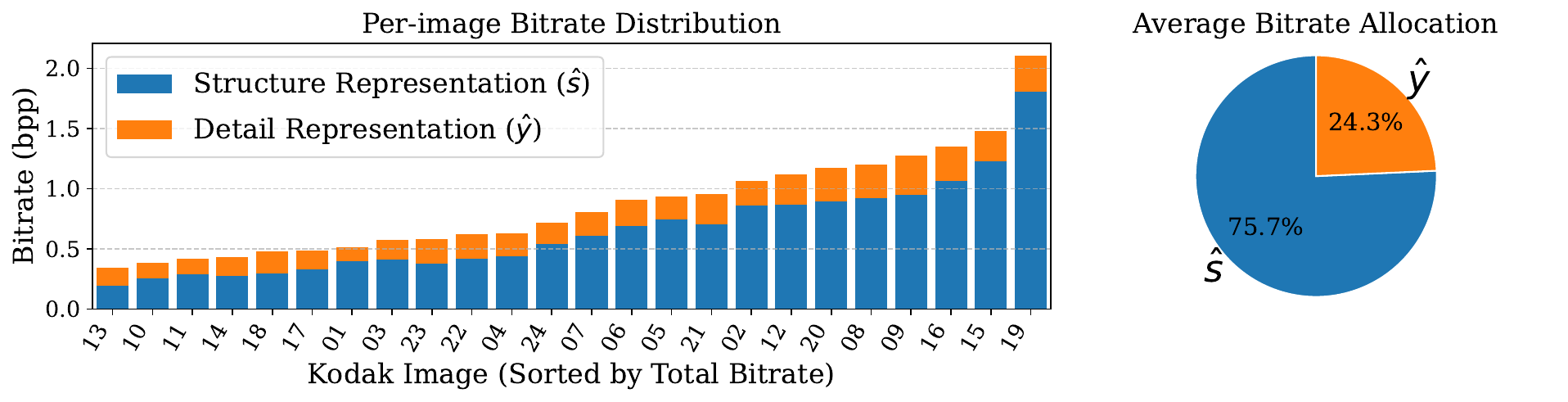}
    \caption{
    The bitrate allocation on Kodak dataset.
    }
    \label{fig:bite}
\end{figure}
\subsection{Energy Distribution Analysis}
\label{sec:energy_analysis}
To empirically verify the hypothesized functional decomposition within our DCIC-sgp framework, we analyze the inter-channel energy distribution of the learned latent representations. This analysis provides insight into the specialized roles learned by the Prior Extractor ($E_s$) and the Conditioned Analysis Transform ($g_a$).

As shown in Figure~\ref{fig:vis_feature} (left), we plot the energy distribution across channels for both the structure prior $s$ and the detail representation $y$. A clear division of labor is observed: the channels of the structure prior $s$ are dominated by high-energy components, whereas the channels of the detail representation $y$ contain predominantly low-energy components. This observation is consistent with the design objective for the Prior Extractor ($E_s$) to capture the high-energy, low-frequency structural information, which in turn allows the Conditioned Analysis Transform ($g_a$) to focus its capacity on the remaining low-energy, high-frequency details.
Furthermore, Figure~\ref{fig:vis_feature} (right) visualizes the channels with maximum energy at different stages of training, highlighting the co-evolutionary behavior of the two networks as they learn their respective, complementary roles.

To empirically validate our functional decomposition, we analyzed the bitrate allocation for our low-bitrate DCIC-sgp-TCM model. 
The analysis presented in Figure~\ref{fig:bite} reveals that the framework prioritizes the image's structural integrity at this low rate point by allocating the majority of the bitrate (approximately 75.7\%) to the structure prior ($\hat{s}$).
This strategic bitrate allocation directly explains the superior visual quality in Figure~\ref{fig:vis}. It allows our model to robustly encode the structural backbone, substantially mitigating the geometric deformation that plagues competing methods at low bitrates.

\begin{figure*}[!t] 
    \centering
    \subfloat[PSNR on ISIC 2019 dataset]{\includegraphics[width=0.485\textwidth]{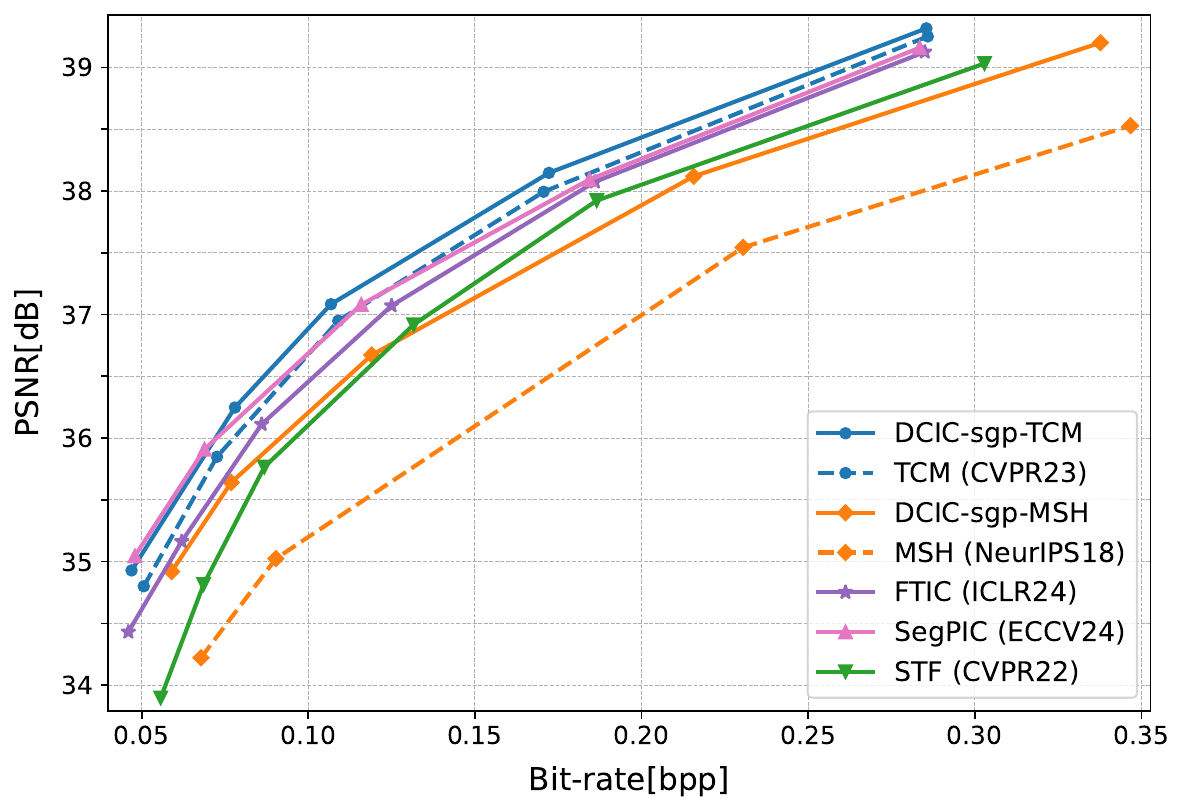}
    \label{fig:ISIC_psnr}}
    \hfill 
    \subfloat[PSNR on Blood Cell dataset.]{\includegraphics[width=0.485\textwidth]{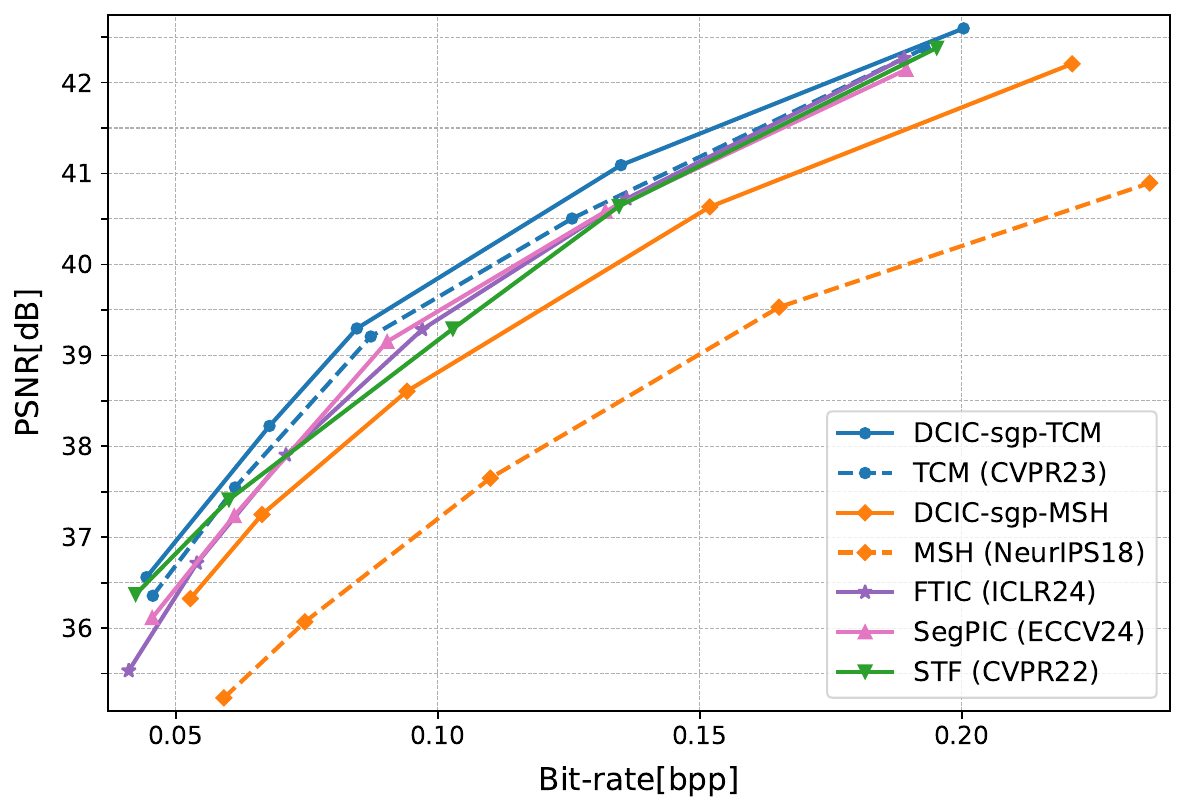}
    \label{fig:Blood_psnr}}

    \caption{Rate-distortion performance (PSNR vs. bpp) of our DCIC-sgp models evaluated on medical datasets. Performance is compared against respective baselines (MSH and TCM) and other leading methods to assess robustness and relative standing in these specific domains.}
    \label{fig:med_performance_comparison}
\end{figure*}
\begin{figure}[!th]
    \centering 
    \includegraphics[trim={30pt 50pt 40pt 50pt}, clip, width=0.95\textwidth]{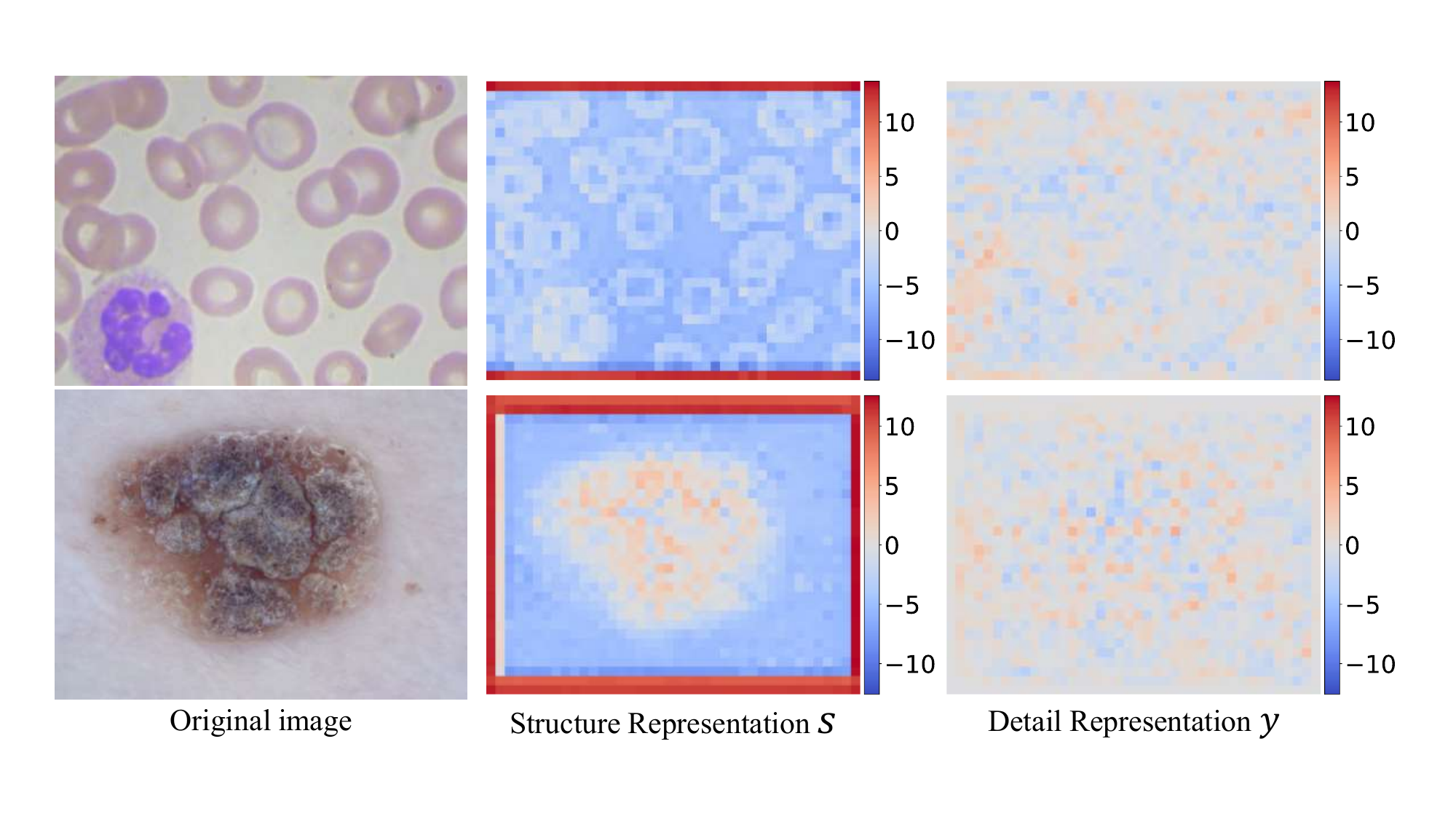}
    \caption{Visualization examples from medical images; Blood Cell (top row) and ISIC (bottom row). Visualizations for the structure prior ($s$) and detail representation ($y$) display the channel with maximum energy. Note that visible edge artifacts in the channel visualizations result from zero-padding applied to the original images to ensure network compatibility. The structure prior $s$ (middle column) captures relevant semantic structures despite the domain shift.}
    \label{fig:med_vis}
\end{figure}
\subsection{Performance on Medical Image Domains} 
To rigorously assess the robustness and generalization capability of our framework beyond the natural image domain used for training, especially concerning the learned structure prior ($s$), we conducted evaluations using our DCIC-sgp models (trained on a subset of ImageNet~\cite{deng2009imagenet}). We selected two medical image datasets representing distinct domains significantly different from natural scenes: ISIC 2019 test images~\cite{isic2019data} (dermatoscopy, for which we used the first 500 images sorted alphabetically from the official test input) and Blood Cell test images~\cite{kaggleBloodCells} (microscopy). 
The rate-distortion performance is presented in Figure~\ref{fig:med_performance_comparison}.
The results demonstrate that our paradigm remains effective on these distinct medical domains, where our DCIC-sgp models consistently outperform their respective baselines (MSH and TCM) across both datasets.
Furthermore, Figure~\ref{fig:med_vis} provides visualizations comparing the original image with the maximum energy channel visualizations of $s$ and the detail representation $y$. The structure prior $s$ successfully captures salient structural elements despite the domain shift.

\section{Conclusion}
In this paper, we introduced Deeply-Conditioned Image Compression with self-generated priors (DCIC-sgp), a new paradigm designed to overcome the limitations of conventional learned codecs. The core of DCIC-sgp is the use of a rich, self-generated structure prior that serves as a powerful prior. Unlike prior art where conditioning is often shallow, we leverage this prior to holistically guide the entire compression pipeline. Most notably, the "deep conditioning" of the analysis transform itself enables a more effective functional decomposition of image information, leading to significant gains in rate-distortion performance. Our work demonstrates the immense potential of creating and deeply integrating powerful internal priors for single-image compression. 
Future work will explore more dynamic, content-adaptive strategies for prior generation and extend the DCIC-sgp paradigm to other modalities like video and 3D data compression.

\section*{Acknowledgments}
This research was supported by the National Natural Science Foundation of China under Grant 62331014 and by Grant 2021JC02X103.


\bibliographystyle{elsarticle-num} 
\bibliography{references}





\end{document}